\documentclass[sigconf,authorversion]{acmart}
\AtBeginDocument{%
  \providecommand\BibTeX{{%
    \normalfont B\kern-0.5em{\scshape i\kern-0.25em b}\kern-0.8em\TeX}}}

\usepackage{graphicx}
\usepackage[]{acronym}
\usepackage{multirow}
\usepackage{array}
\usepackage{makecell}
\usepackage{subfigure}
\usepackage{caption}
\usepackage[normalem]{ulem}
\usepackage[textsize=footnotesize]{todonotes}  

\acrodef{API}{Application Programming Interface}
\acrodef{AUC}{Area Under the Curve}

\acrodef{DB}{Database}

\acrodef{FOR}{false omission rate}
\acrodef{FPR}{false positive rate}

\acrodef{GDPR}{General Data Protection Regulation}

\acrodef{HDFS}{Hadoop Distributed File System}
\acrodef{HAIDS}{Human-AI Decision-making System}

\acrodef{IRS}{Interface Review System}
\acrodef{IHAIRS}{Interface Human-AI Review System}

\acrodef{LIME}{Local Interpretable Model-Agnostic Explanations}

\acrodef{ML}{Machine Learning}
\acrodef{MLDS}{ML Decision System}

\acrodef{OEC}{Overall Evaluation Criterion}
\acrodef{OOP}{Object Oriented Programming}

\acrodef{RCT}{Randomized Control Trial}
\acrodef{RDBMS}{Relational Database Management System}
\acrodef{REST}{Representational State Transfer}

\acrodef{SQL}{Structured Query Language}
\acrodef{SE}{Standard Error}

\acrodef{XAI}{Explainable AI}

\copyrightyear{2021}
\acmYear{2021}
\setcopyright{acmlicensed}
\acmConference[FAccT '21]{Conference on Fairness,
Accountability, and Transparency}{March 3--10, 2021}{Virtual Event, Canada}
\acmBooktitle{Conference on Fairness, Accountability, and Transparency (FAccT
'21), March 3--10, 2021, Virtual Event, Canada}
\acmPrice{15.00}
\acmDOI{10.1145/3442188.3445941}
\acmISBN{978-1-4503-8309-7/21/03}

\copyrightyear{2021}
\acmYear{2021}
\setcopyright{acmlicensed}\acmConference[FAccT '21]{Conference on Fairness,
Accountability, and Transparency}{March 3--10, 2021}{Virtual Event, Canada}
\acmBooktitle{Conference on Fairness, Accountability, and Transparency (FAccT
'21), March 3--10, 2021, Virtual Event, Canada}
\acmPrice{15.00}
\acmDOI{10.1145/3442188.3445941}
\acmISBN{978-1-4503-8309-7/21/03}

\begin{document}

\title{How can I choose an explainer? An Application-grounded Evaluation of Post-hoc Explanations} 


\author{Sérgio Jesus}
\affiliation{%
  \institution{Feedzai, DCC-FCUP\\ Universidade do Porto}
}
\email{sergio.jesus@feedzai.com}

\author{Catarina Belém}
\affiliation{%
  \institution{Feedzai}
}
\email{catarina.belem@feedzai.com}

\author{Vladimir Balayan}
\affiliation{%
  \institution{Feedzai}
}
\email{vladimir.balayan@feedzai.com}

\author{João Bento}
\affiliation{%
  \institution{Feedzai}
}
\email{joao.bento@feedzai.com}

\author{Pedro Saleiro}
\affiliation{%
  \institution{Feedzai}
}
\email{pedro.saleiro@feedzai.com}

\author{Pedro Bizarro}
\affiliation{%
  \institution{Feedzai}
}
\email{pedro.bizarro@feedzai.com}

\author{João Gama}
\affiliation{%
  \institution{LIAAD, INESCTEC\\Universidade do Porto}
}
\email{jgama@fep.up.pt}



\begin{abstract}
There have been several research works proposing new Explainable AI (XAI) methods designed to generate model explanations having specific properties, or desiderata, such as fidelity, robustness, or human-interpretability. However, explanations are seldom evaluated based on their true practical impact on decision-making tasks. Without that assessment, explanations might be chosen that, in fact, hurt the overall performance of the combined system of ML model + end-users. This study aims to bridge this gap by proposing XAI Test, an application-grounded evaluation methodology tailored to isolate the impact of providing the end-user with different levels of information. We conducted an experiment following XAI Test to evaluate three popular post-hoc explanation methods -- LIME, SHAP, and TreeInterpreter -- on a real-world fraud detection task, with real data, a deployed ML model, and fraud analysts. During the experiment, we gradually increased the information provided to the fraud analysts in three stages: \textit{Data Only}, \textit{i.e.}, just transaction data without access to model score nor explanations, \textit{Data + ML Model Score}, and \textit{Data + ML Model Score + Explanations}. Using strong statistical analysis, we show that, in general, these popular explainers have a worse impact than desired. Some of the conclusion highlights include: i) showing \textit{Data Only} results in the highest decision accuracy and the slowest decision time among all variants tested, ii) all the explainers improve accuracy over the \textit{Data + ML Model Score} variant but still result in lower accuracy when compared with \textit{Data Only}; iii) LIME was the least preferred by users, probably due to its substantially lower variability of explanations from case to case.

\end{abstract}


\maketitle

\section{Introduction}

\begin{figure}[h]
  \includegraphics[width=\linewidth]{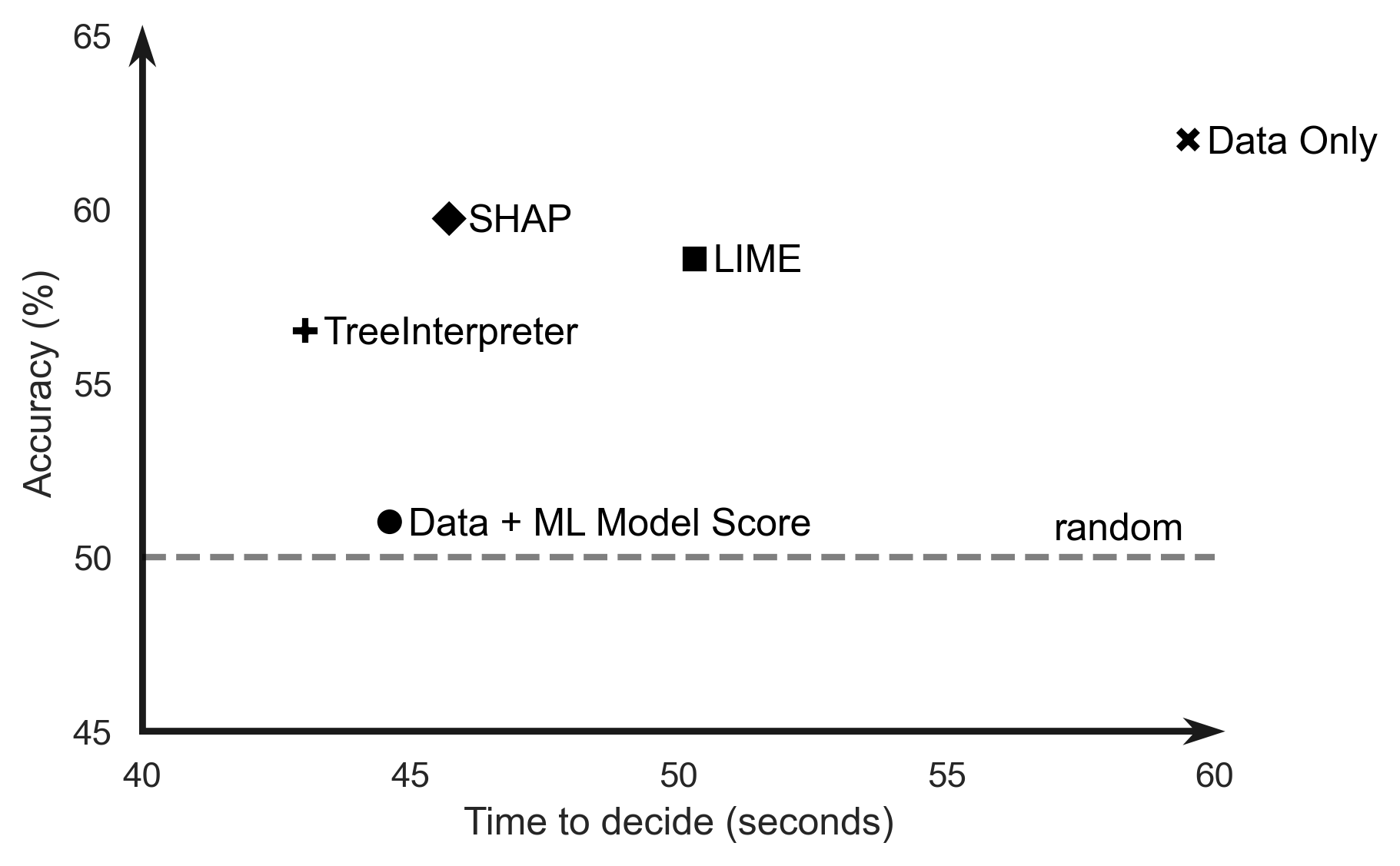}
  \caption{End-users' average decision accuracy \textit{vs.} average time to make a decision for each variant tested in our evaluation experiment of \textit{post-hoc} explanations. We used balanced samples of positive and negative instances, therefore, a random decision process would have $50\%$ accuracy. }
  \label{fig:intro_fig}
\end{figure}
\subsection{The evaluation problem in Explainable AI}
The interest in ML models' explainability has been growing in the last years, as a counteractive effort to the current AI black-box paradigm, coupled with increased public scrutiny and evolving regulatory law \cite{GDPR,ribeiro2016lime,lundberg2017,buolamwini2018gender}. However, this growth in \ac{XAI} research work has not been accompanied by effective evaluation methodologies~\cite{doshivelez2017rigorous}. The field is still in its early stages.

Even though every \textit{persona} interacting with a black-box ML model may benefit of model explainability, each \textit{persona} has a specific role, objectives, actions at disposal, background, domain knowledge, and, consequently, different explainability requirements~\cite{tomsett2018interpretable,mohseni2018multidisciplinary,amarasinghe2020explainable}. As a result, the evaluation of \ac{XAI} methods must be performed with the target persona and the associated task in mind ~\cite{mohseni2018multidisciplinary,amarasinghe2020explainable}. 
 Notwithstanding, in seminal works of \ac{XAI} methods, it is common to see introduced one or multiple \textit{ad-hoc} evaluation setups, mostly focused on ideal explanations desiderata \cite{Ribeiro:2016, lundberg2020, anchors:aaai18, deeplift:shrikumar2017}. In some cases, user experiments are simulated \cite{Ribeiro:2016} or even completely discarded from the evaluation step \cite{lundberg2018tree}.  As a consequence, there is a lack of systematic comparison between different methods accurately and exhaustively. These reasons culminate, ultimately, in general skepticism about the reliability and usefulness of \ac{XAI} methods, especially when the application is of high responsibility. 

\subsection{The impact of showing explanations}
In this work, we focus on \ac{XAI} evaluation having the end-user as target \textit{persona}. We consider the end-user as the decision-maker, the human-in-the-loop, who usually is a domain expert, such as a judge, a doctor, or a fraud analyst. We argue that, for end-users, the value of explanations is heavily determined by how useful they are to the associated decision task and, for that reason, that their evaluation should be made by measuring their impact in the performance of the end-users. This implies involving end-users in the evaluation process, in a setup with a real task and real data. Additionally, metrics should reflect directly the users' performance, \textit{e.g.}, how accurate the decisions are, or how fast they are made.

We propose XAI Test, an application-grounded evaluation methodology tailored to isolate the impact of gradually providing different levels of information to the end-user. A useful \ac{XAI} method produces explanations that improve the overall performance of the combined system of ML model + end-user. To perform a reliable assessment, XAI Test requires testing different combinations of data, model score, and \ac{XAI} methods in a real task with real end-users. Specific performance metrics must be defined (\textit{e.g.}, accuracy or decision time), the agreement between end-users is considered on each variant, and user perception captured through questionnaires. Lastly, statistical tests are employed to detect significant differences between each variant.

\subsection{The experiment}

Using XAI Test, we conducted an empirical evaluation in the task of fraud detection in financial transactions. We employed three different \textit{post-hoc} explainers and observed their impact on human-in-the-loop performance, measuring accuracy, recall, \ac{FPR}, and decision time. We additionally collected the users' perception of usefulness, variety, and relevance of each presented explanation. 

We quantified and isolated the impact of the different interacting parties in a Human-AI collaborative setting by following a three-stage evaluation approach with increased information. Figure \ref{fig:intro_fig} shows how the average accuracy of the decision varies with the decision time for each of the evaluated variants. We observe a clear trade-off between effectiveness and efficiency as the end-user gets access to additional ML model information. In particular, we observe that when no model-related information is shown to the end-user (\textit{i.e.,} \textit{Data Only}), although slower, leads to more accurate decisions. Conversely, the accuracy obtained in the mid-level information stage (\textit{Data + ML Model Score}) yields faster decisions but much worse accuracy - a result that is partially improved by adding model explanations.

%
%
\section{Related Work}
\label{sec:related_work}

In this section, we provide an overview of the current evaluation paradigm in \ac{XAI} research. In particular, we briefly discuss the often considered desiderata, as well as the different techniques used to measure them. We end by enumerating a few representative state-of-the-art evaluation approaches and by describing how these fail to convey a robust analysis of the real impact of \ac{XAI} methods in real-life Human-AI decision-making systems. 

\subsection{Desiderata}

Most research work on \ac{XAI} measures some kind of \textit{proxy} of intuitive desiderata for the ideal explanation, such as \textbf{fidelity} or faithfulness \cite{Ribeiro:2016, PlumbMT18:MAPLE}, which states that surrogate models that are used to obtain \textit{post-hoc} explanations should be able to mimic the behavior of the explained ML model; \textbf{robustness} or stability \cite{AlvarezMelis2018}, which measures whether similar input instances get similar explanations; \textbf{human-interpretability} or comprehensibility \cite{narayanan2018}, which measures how easily a human interprets the result from the explanation method.

Despite being common sense that a good explanation must have high fidelity, be robust, and be intelligible, those characteristics by themselves do not say much about the actual benefit of having an explanation in a specific real-world application, nor do the measurements completely represent those characteristics.

Previous work often assumes that a model is \textbf{interpretable} because it belongs to a certain family of models -- such as sparse linear models,
decision trees, and rules lists \cite{Cohen:1999, friedman2008, Dembczynski2008, lakkaraju2016}, or additive models \cite{Vaughan2018, Zhang:2019, caruana2015intelligible} -- and the only focus when generating explanations is on the accuracy of those models. These explanations are directly derived from interpreting the ML model parameters. 
Most of the times, these over-simplified definitions of model intelligibility are detached from the requirements of real-world applications \cite{lipton2016mythos}. In general, these simpler models have much lower predictive accuracy than other more complex models, such as deep neural networks or tree ensembles. 
Only in a few high-stakes tasks (such as credit scoring \cite{Cynthia2019}) is the complexity of an ML model viewed as an actual limitation, and only in these particular cases, there is no alternative to simpler, more intelligible models.

Several works assess \textbf{fidelity} as a measure of the quality of an explanation. Fidelity has been assessed both directly \cite{PlumbMT18:MAPLE, Ribeiro:2016, sanchez15:towards, TanCHL18}, by measuring differences in predictions of the surrogate and explained models, as well as indirectly \cite{anchors:aaai18}, by measuring how well a human can predict the output of a ML system with and without being exposed to explanations. Again, this is another metric detached from real-world impact of showing an explanation to a given persona, as it focuses on how well an \ac{XAI} model approximates the function learned by the original ML model.

Other works defend the importance of \textbf{robustness}. It is measured by directly computing how much the output of an explanation method changes with its input \cite{MelisNIPS2018, AlvarezMeli2018} or by showing the sensitivity of explanations to adversarial attacks \cite{GhorbaniAZ19}. 
However, these metrics are not directly related to how an explanation might help the end-user to better perform their task.

\textbf{Interpretability} is also assessed by measuring how approximate a \ac{XAI} method explanation is to an explanation produced by a human expert \cite{lundberg2017, Kim2018_TCAV}. 
Those approaches are somehow restricted to tasks where the behavior of humans is intuitive, and generally close to the ground truth (such as problems in natural language processing and computer vision), but may not be suitable to complex predictive tasks based on tabular data, where the analysis has to take into account multiple features and interactions, making the task harder and less intuitive.

The way \ac{XAI} desiderata is being interpreted and measured is disperse and lacking in consensus, as shown by the different methods to measure the same property. Several problems are pointed to current practices, such as non-overlapping and discordant motivations and objectives for interpretability \cite{lipton2016mythos}, attributing the same level of interpretability to ML models originating from the same model class \cite{doshi2017towards}, or the lack of evaluation of \ac{XAI} methods with the intended end-users \cite{asylum}.

Frameworks have been developed \cite{mohseni2018multidisciplinary, Murdoch_2019} as an attempt on tackling the challenges of \ac{XAI} evaluation, however, these frameworks are still recent and have yet to see wide adoption.
The field is missing a systematic and objective way of comparing explanation methods \cite{lipton2017doctor, schmidt2019quantifying}, which promotes research practices where each work uses customised metrics and desiderata that are thought to be the most adequate, encumbering the choice of \ac{XAI} methods for a given task. This is especially important in scenarios of real-world Human-AI decision-making systems, where \ac{XAI} methods may have a greater impact. 

\subsection{Evaluation Practices}
\label{ssection:xai_eval}
While many \textit{ad-hoc} evaluation setups have been used to empirically validate research on XAI methods, these either found on idyllic desiderata or overlook the human-in-the-loop and their explainability needs. In an attempt to standardize the existing XAI evaluation approaches, Doshi-Velez and Kim \cite{doshivelez2017rigorous} propose a taxonomy to categorize the different types of XAI evaluation practices. In their work, the authors subdivide the evaluation practices into three distinct groups, depending on whether it resorts to humans or not and on the task they are being employed on. The first group encompasses automated evaluation on proxy tasks and is designated as \textit{functionality-grounded} evaluation. Experiments in this category may try to simulate human behavior \cite{anchors:aaai18}, and apply these simulations to real tasks, such as fraud detection \cite{weerts2019}. Other works do not consider the human factor as part of the evaluation \cite{PlumbMT18:MAPLE, Ribeiro:2016}. 



Both other groups of evaluation methods use humans in the process of evaluation but differ on the task being done. If the evaluation task is a simplified proxy of a real task, the method is designated \textit{human-grounded} evaluation, while if the task is in a real-world setting, the method is deemed \textit{application-grounded} evaluation. These methods introduce the human component in the evaluation loop to collect feedback in the form of questionnaires, surveys, interviews, performance at the task, among others. Their focus, however, shifts from how humans perceive and interact with the explanations in \textit{human-grounded} evaluation, to how it affects the whole system performance in \textit{application-grounded} evaluation.

The evaluation of explanations through experimentation has been done in several past works. Most experimental studies use proxy tasks with real human subjects, \textit{i.e.}, human-grounded experiments, such as trivia answer \cite{shi2019}, clinical prescription simulation \cite{narayanan2018, lakkaraju2016}, detection of deceptive reviews \cite{vivian2019}, comparison between human feedback and explainer output \cite{lundberg2017}, or human prediction of model output on unseen instances based on the explanation of the model behavior \cite{anchors:aaai18}.

By analysing the experiments conducted in other works, there is a clear gap in evaluation using real tasks with real end-users. More often than not, explanations are employed in mocked tasks, and the results obtained can not be generalized to high responsibility real-world tasks. Simulating human behavior is prone to human bias, since in many cases it depends on the developers' own intuition of the problem, and may diverge from reality, producing unrealistic results. Additionally, seldom do these experiments compare explanation methods, but rather test different visualizations or output types for these methods, which emphasizes more on the presentation rather than explanations' content. 

%

%
%
\section{Evaluation Methodology}
\label{sec:methodology}

The evaluation of the true impact of a given explanation in the end-user experience is not an easy task. 
Ideally, it should be focused on objectively measuring its utility (or usefulness) in the users' decision making process. 
This should rely on the collection of metrics from real users while performing real tasks on real data.

We propose \textbf{XAI Test}, an application-grounded evaluation me\-thodology that relies on realistic settings and statistical tests to robustly assess and compare the explanations' utility of different \ac{XAI} methods, using metrics that correspond to the performance of the user.
Rather than evaluating explainability through idyllic desiderata, we opt for evaluating it through metrics that quantify the true impact in the human decision-making. 

The methodology consists of the following steps: (1) formulate the hypotheses; (2) outline the experimental setup; (3) define the statistical tests to report the results with; (4) conduct the three stages of the experiment; and (5) apply statistical tests to obtained measurements.

With this methodology, we aim to find answers to a set hypotheses (\textit{e.g., is method $A$ more efficient than method $B$? Is it more accurate?}). In the case of an \ac{XAI} experiment, these hypotheses are related to the utility of the explanations and how they impact the end result of a given task.
To support or reject the formulated hypotheses, it is necessary to objectively measure users' performance at the task (\textit{e.g.}, through accuracy, or decision time). 
It is also important do define other elements of the experiment, including the explainers, ML models, corresponding configurations to test, number of users that partake on the experiment, datasets, and other task-specific details, such as experiment scheduling and used software. 
Equally important for ensuring a robust evaluation is the confidence of the reported results. To this end, we define the appropriate statistical tests as well as their parameters, which are significance level, statistical power, effect size, and sample size. A prior knowledge of the distributions is required to choose these parameters. In Section \ref{ssec:stats_tests}, we elaborate on the choices made in terms of hypothesis testing.
The ensuing step is then to conduct the experiments in a way that isolates the impact of explanations in the decision making process. For this reason, we advocate for the execution of, at least, three stages, each providing added levels of information: (1) \textit{Data only}, (2) \textit{Data + ML Model Score}, and (3) \textit{Data + ML Model Score + Explanations}. 
Finally, the last step of the proposed methodology concerns the collected results and their analysis. 

The following sections describe the methodology employed in the evaluation of \ac{XAI} methods. This includes the way explanations are employed, the measured metrics, and the battery of statistical tests to determine any significant difference.

\subsection{Metrics Choice}
\label{ssec:metrics}

The metrics choice is task-specific. 
In Human-AI cooperative systems where the true data labels are known, it is possible to combine this information with the user decision to compute performance metrics (based on the confusion matrix), such as recall, \ac{FPR}, precision, or \ac{FOR}. 
These measures allow us to objectively quantify the impact of different components (\textit{e.g.}, model score and/or different explanation types) in the human decision-making process. 
In practice, accuracy, recall, and \ac{FPR} are better choices, because the denominator either depends on the sample size, or on the number of label positives and label negatives of the sample. Since these are constant over the course of the experiment and do not depend on the number of predicted positives and negatives (as it is the case for a metric such as precision and \ac{FOR}), we can determine \textit{a priori} the exact sample size for each metric.

In most systems, time is also a determining factor and should, therefore, be monitored during system modifications. In Human-AI decision making systems, explanations serve to help the human-in-the-loop  to make a faster decision, by pointing them to what the model perceives to be the most important information for the decision. Consequently, this is an important aspect to measure when discerning the impact of explanations in decision-making processes.

\begin{figure}[hbtp]
\caption{Questionnaire performed to the users after each instance with explanation.}
\includegraphics[width=\linewidth]{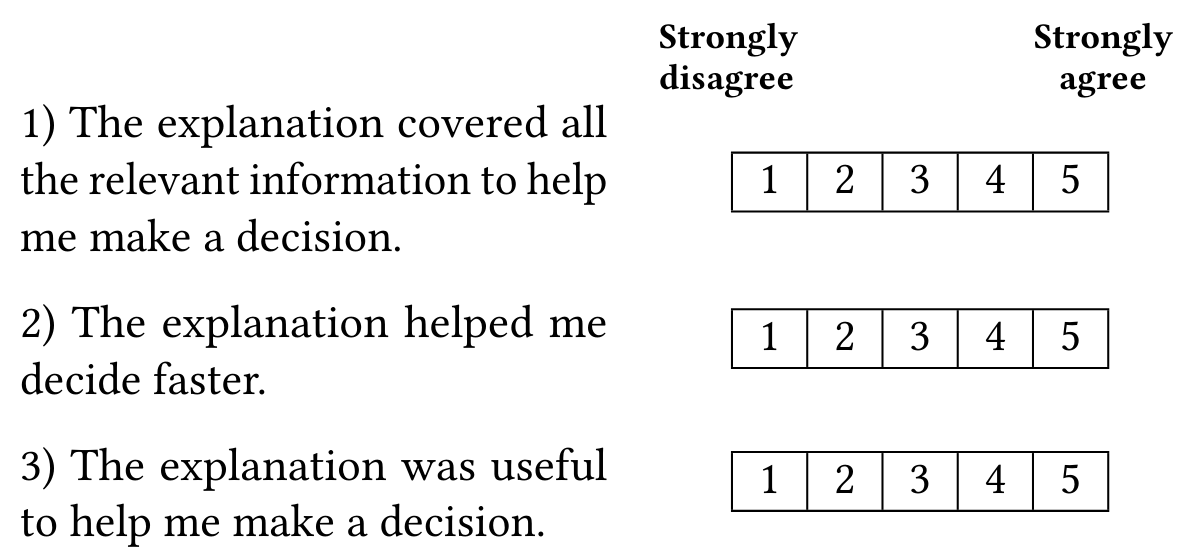}
\label{fig:Questionnaire}
\end{figure}

Another relevant point, despite being more subjective, is the user's perception of the explanation quality, including its relevance and usefulness. For this reason, we propose a set of predefined five-point \textit{Likert-type} scale questions, specified in Figure \ref{fig:Questionnaire}.

Finally, often times, decisions diverge from user to user. We expect the addition of more information (\textit{e.g.}, model scores and/or explanations) to mitigate such differences. To accurately measure this effect, we use an agreement set where a subset of the data is shared between users with the intent of computing the metrics of agreement. We use \textit{Fleiss' Kappa} \cite{fleiss1973} as the agreement metric because our experiments will incorporate multiple users. Additionally, we calculate the average agreement, which is the average pair-wise agreement between users.

\subsection{Experimental Stages}
\label{ssec:experiment_types}

While, in the first stage of the experiment, humans only have access to instance-specific information (feature data), in the second the human is provided with information of the model score, calibrated for simplification. Consequently, users may sometimes perceive it as a measure of \textit{how confident the model is about predicting a given class}: scores closer to 1 or 0 express confidence, whereas scores around 0.5 convey more uncertainty.  

The third stage of the experiment involves, in addition to the ML model score, the explanations. How and which information to show for which explainer should be defined in the experimental setup. There are many degrees of freedom when configuring an explainer: the explainer type (\textit{e.g.}, self-explainable, \textit{post-hoc}), the number of features to consider, how to represent the explanation (\textit{e.g.}, feature contributions, heatmaps, scores, visualizations) as to minimize the cognitive load during the task execution. Another important aspect to pay attention to are the biases that may arise if explanation methods are distinguishable due to some factor (\textit{e.g.}, their representation). Mitigating their representational differences is, therefore, a preventive step towards isolating the quality and relevance of the explanation methods from all the other possible visual factors. 

\subsection{Statistical Tests}
\label{ssec:stats_tests}

The appropriate choice of a statistical test depends on two factors: (1) the metric distribution and (2) the end-goal of the test. 
Most statistical tests aim at identifying significant differences between measured averages of performance metrics in different scenarios (control \textit{vs} treatment). In this case, we use of \textit{Chi-squared} test \cite{pearson1900} for multiple group comparison of instance-level binary metrics, such as accuracy, recall, and \ac{FPR}.
Conversely, for continuous performance metrics like decision time, we use a non-parametric test named \textit{Kruskal-Wallis} \textit{H} \cite{kruskal1952} to validate whether the samples belong to the same underlying process. This test is particularly suited for non-normal distribution of continuous variables. 

We are interested in comparing pairs of groups and, specifically, in running comparisons between each variant and the control group. In these cases, we use \textit{Chi-squared} test with the pairs to be tested in the performance values, and the Mann-Whitney \textit{U} test \cite{Mann1947} on continuous data. P-values must also be corrected for family-wise error rate with the \textit{Holm-Bonferroni} method \cite{holm1978}.

In order to quantify the perceived usefulness and relevance of the explanations measured through the questionnaire, we aim to identify distribution differences between different explainers for the proposed questions. We find the \textit{Kruskal-Wallis} \textit{H} to better suit this goal when comparing multiple variants. To report the results of paired tests, we apply the \textit{Kolmogorov-Smirnov} test \cite{kolmogorov1933} corrected with the \textit{Holm-Bonferroni} method \cite{holm_1979}.


%
%
\section{Experiments} 
\label{sec:experimental_setup}

We employ our proposed application-grounded methodology, XAI Test, to evaluate and compare different explanation methods in a real-world decision-making task: fraud detection in payment transactions. We had access to a real fraud prevention system comprising a deployed ML model that predicts the risk of fraud for each payment transaction in a given online retailer. The fraud analyst is responsible for accepting or declining payment transactions for which the ML model is more uncertain about (the score is within a review band). This decision-making task is performed through a web interface in which the fraud analyst can inspect details of the payment transaction (\textit{e.g.}, shipping address, billing email, time since last transaction) which represents the feature data (\textit{i.e.}, \textit{Data Only}) together with the risk score, given by the ML model, and an explanation. 

While business requirements aim for more effective and efficient decisions, often, the model information is not sufficient to meet such criteria (\textit{e.g.}, disagreement between fraud analysts and ML model or even mistrust in the model predictions). In an attempt to bridge this Human-AI gap, we conjecture that \textit{explanations promote better human performance in such predictive fraud task}. Therefore, the prime goal of this experiment is to assess the real impact of showing explanations to real humans (the fraud analysts) interacting with a real ML model. 


\subsection{Experimental Hypotheses}
\label{ssec:experimental_hypoehtess}

As the first step of XAI Test, we formulated our hypotheses. Since we used a production system without permission to modify the ML model, we focus on the evaluation of \textit{post-hoc} explanation\footnote{Explanations produced by post-hoc methods.}  methods. With this in mind, we set out to answer the following hypotheses: 

\begin{itemize}
    \item \textit{H1}. Showing fraud analysts the \textit{ML Model Score} improves their performance \footnote{Defined in Section \ref{ssec:experimental_setup}.} over \textit{Data Only};
    \item \textit{H2}. Showing \textit{post-hoc} explanations significantly improves human performance over \textit{Data Only} and/or \textit{Data + ML Model Score};
    \item \text{H3}. Explanations from different \textit{post-hoc} explainers impact humans differently; Assuming that humans trust the explanations, some explainers promote more effective and/or efficient decisions;
    \item \textit{H4}. Each \textit{post-hoc} explainer is perceived differently in terms of relevance, usefulness, and diversity;
    \item \textit{H5}. Showing explanations increases fraud analysts agreement over the same set of transactions;
    \item \textit{H6}. Showing model score information increases fraud analysts agreement over the same set of transactions.

\end{itemize}

\subsection{Experimental Setup}
\label{ssec:experimental_setup}

We evaluate the above hypotheses using metrics indicative of the fraud analysts' performance in terms of both efficiency and efficacy at the decision-making task. 

\textbf{Metrics:}
    We use the average decision time (of fraud analysts) as an efficiency measure and we use accuracy, \ac{FPR}, and recall as measures of their effectiveness.
Moreover, to address \textit{H4}, we also measure their perceived relevance, usefulness, and diversity of the explanations through the questionnaire in Figure \ref{fig:Questionnaire}.

\textbf{ML model}:
As an application-grounded evaluation of a real-world system, we used the fraud prevention system's ML model: a Random Forest's variant \cite{breiman2001}.

\textbf{Explainers}:
Among the various \ac{XAI} methods for tabular data, we opted for two of the most commonly used \textit{post-hoc explainers}: LIME \cite{Ribeiro:2016} and SHAP \cite{lundberg2017}. 
In particular, we leveraged the fact that the model is a decision tree ensemble to use the tree-based SHAP explainer
- TreeSHAP \cite{lundberg2018tree}.
We also included a third explainer specifically tailored for tree-based algorithms, known by ML practitioners as TreeInterpreter \cite{saabas2015treeinterpreter}. 
%
In terms of hyperparameters, we ran a few sensitivity tests to determine the most appropriate hyperparameters for the proposed task. From this analysis, we concluded that both SHAP and TreeInterpreter could be used with their out-of-the-box parametrization, whereas LIME had to be tweaked, specially, due to its stochastic nature\footnote{LIME's internal local fidelity metric showed improvements exclusively upon variations on the number of perturbed samples.}. Thus, besides the random seed, we also set the number of perturbed samples to 5k.

\textbf{Explanation format}:
The explanations format for the three explainers consists of pairs of \textit{feature-contribution}. We decided to only display the top $6$ pairs based on contribution value. Unlike other tabular explanation formats, such as decision lists and decision sets \cite{lakkaraju2016}) the \textit{feature-contribution} format benefits from its readability, simplicity and visualization flexibility.

Furthermore, to create a seamless experiment, we used this output's simplicity to homogenize the explanations representation across explainers. Given a set of feature-contribution pairs, we: (1) sort it in descending order by absolute contribution value\footnote{Higher contributions reflect more important features.}, and (2) transform it into a human-readable format. This transformation comprises mapping the feature name to a natural language description plus parsing the feature value (\textit{e.g.}, converting time from seconds to days).

\begin{figure}[hbtp]
  \includegraphics[width=\linewidth]{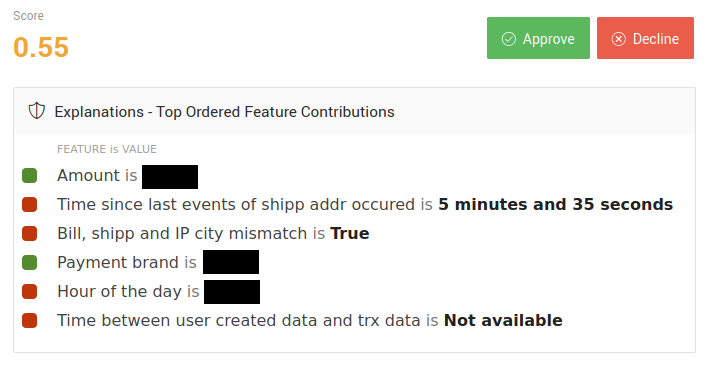}
  \caption{Visual representation of an explanation, as viewed by fraud analysts during the experiment (obfuscated to preserve privacy).}
  \label{fig:exp_example}
\end{figure}

We further added a color-based visual cue to reflect the changes in the associated suspicious risk (score): negative contributions represented with \textit{green}, as they contribute for lower scores and consequently legitimate transactions, and, conversely, positive contributions represented with \textit{red}. Figure \ref{fig:exp_example}, illustrates an explanation shown to a fraud analyst during the experiments.


\textbf{Users}: 
Three professional fraud analysts partook in the experiment. They were all experienced users of the fraud detection system used in the experiment.

\textbf{Data}:
Two different samples were considered: (1) a training sample, derived from the same data set used to train the ML model, and (2) an experiment sample, from the production period of the ML model. We used the former as the background for LIME (to obtain information about features distributions). To create it, we randomly sampled 100k transactions from the model's training set. Conversely, the sample for running the experiment itself, dubbed experiment sample, was extracted from the model's production period  (November 2019), for which we had fraud labels. We extracted a stratified sample to attain $50\%$ fraud prevalence.

To replicate a real scenario for the experiment the sample exclusively comprises transactions that lie in the review band, \textit{i.e.}, transactions with higher model uncertainty.
The final experiment sample size totals $1300$ transactions. In the following section, we disclose how these transactions were distributed across the different experiment stages.

\subsection{Experiment Outline}
\label{ssec:task_and_participants}
\begin{table*}[thbp]
  \captionsetup{width=.9\textwidth}
  \caption{
  Performance, time and agreement metrics for all variants of the experiment. Statistical significance is tested between each explainer and each of the two groups that do not show explanations or only among explainers.  
  $^{\star}$ indicates significant difference with Data Only; no statistically significant difference was detected between each explainer and the Data + ML Model Score; $^{\blacklozenge}$ indicates significant difference with all other explainers. The agreement metric is \textit{Fleiss' Kappa}.}
  \label{tab:metrics}

\begin{tabular}{c c c l l l l l}
    \hline
        \multirow{2}{*}{\textbf{Group}} & \multirow{2}{*}{\textbf{Explainer}} & \multirow{2}{*}{\textbf{Sample Size}} & \multicolumn{4}{c}{\textbf{Metrics}} & \multirow{2}{*}{\makecell{\textbf{Agreement}}} \\
    
    \cline{4-7}
     & & & \textbf{accuracy (\%)} & \textbf{recall (\%)} & \textbf{FPR (\%)} &  \textbf{time (s)} \\
    \hline
   
       \makecell{Data Only}  & - & 200 & \makecell{\textbf{62.00}} & \makecell{\textbf{35.87}} & \makecell{15.74} & \makecell{59.50} & \makecell{0.41} \\
    
    \makecell{Data + \\ ML Model Score} & - & 200 & \makecell{51.02$^{\star}$} & \makecell{25.00} & \makecell{19.57} &  \makecell{44.61$^{\star}$} & \makecell{-0.02} \\
    
      \multirow{4}{*}{\makecell{Data + \\ ML Model Score + \\ Explanations}} & LIME & 300 & \makecell{58.59} & \makecell{27.03} & \makecell{\textbf{10.07}} &  \makecell{50.29$^{\star}$} & \makecell{\textbf{0.53}} \\
    
      & TreeInterpreter & 300 & \makecell{56.52} & \makecell{25.55} & \makecell{12.67} &  \makecell{\textbf{43.03}$^{\star\blacklozenge}$} & \makecell{0.30} \\
     
    & SHAP & 300 & \makecell{59.73} & \makecell{31.08} & \makecell{12.00} &  \makecell{45.72$^{\star}$} & \makecell{0.15} \\
    \hline
\end{tabular}
\end{table*}

We conducted all three stages of the experiment, as XAI Test suggests (see Section \ref{ssec:experiment_types}). Given that each stage added levels of information, we decided to run them in a way that allows fraud analysts to incrementally stabilize their mental model of the task (as they adapt to new information within the system). This leads to the following experiment outline:

\begin{enumerate}
    \item \textit{Data only}: information exclusively about the transaction (payment details and history) is available;
    \item \textit{Data + ML Model Score}: both transaction data and the model score are available;
    \item \textit{Data + ML Model Score + Explanations}: all of the above information is complemented with an explanation (from LIME, SHAP, or TreeInterpreter) of the model score. 
\end{enumerate}

As our baseline, we considered the stage where every information except the data was withheld from fraud analysts (\textit{Data only} stage), as it allows us to isolate and quantify the real impacts of different information types in the human's performance in (and understanding of) the task. 
In the absence of prior knowledge about the metrics distribution for this particular task, we used a total of $400$ transactions to conduct the two initial stages of the experiment ($200$ for each stage). Each of these samples were created without replacement from the experiment sample (stratified by fraud label to keep fraud prevalence at $50$\%). We found $200$ transactions to be a good compromise between the pressing time and business constraints (\textit{e.g.}, availability of the analysts) and the quality and rigor of the experiment.

On the other hand, we leveraged the results obtained in the initial experiments (\textit{Data only} and \textit{Data + ML Model Score}) to compute the sample size required to obtain significant results at the desired power, $\beta$, significance level, $\alpha$, and effect size, $\delta$. 
We set $\delta=15$ because we found it to be a good compromise between sample size and the minimum difference detection. Moreover, we defined $\beta=1-\alpha=0.9$, since we perceive both error types associated with statistical hypothesis testing (type I and type II) to be of equal importance during the experiment. In the end, and assuming the proxy estimates of the analysts' distribution were representative of their true performance, we concluded that a sample with $300$ transactions would suffice for rigorously running the third stage of the experiment, the \textit{Data + ML Model Score + Explanations} stage. Each sample was divided equally for each analyst. Each analyst reviewed the same number of transactions for every explainer in the experiment ($100$ transactions per explainer), which guaranteed the results were equally balanced and that the experiment results were not skewed towards a specific explainer or user. 

To address hypothesis \textit{H5}, we defined a subset of each sample to belong to an agreement set. In practice, this implies that all users reviewed the same exact transactions of the agreement set. This set accounted for about $12.5\%$ of the transactions on every experiment stage.

%
%
\section{Results and Discussion}
\label{sec:results}

In this section, we evaluate how various levels of information affect the human's decision-making process in a fraud detection task. 
We first examine the impact of disclosing information about the ML model score when compared to withholding that information. 
We also analyse the impact of showing different \textit{post-hoc} explanations on top of the information about the ML model score. We discuss the obtained results in terms of human effectiveness and efficiency at detecting fraudulent transactions. 

Table \ref{tab:metrics} shows the experiment results for the conducted three-stage experiment (each stage reflects a group). Besides isolating the contributions of the different system components, this table also comprises the evaluation results of three popular \textit{post-hoc} explanation methods, being one of the most comprehensive evaluation and comparison of \ac{XAI} methods to date.  

Our results show that data alone induces better decisions, while showing model scores or model scores with explanations significantly improves the decision time. Our results suggest that, in practical settings where decision speed is a main requirement, ML models explanations carry a significant speed up in human decision-making, as depicted in Figure \ref{fig:times}. Additionally, data alone carries a better result in both accuracy and recall, registering even a significant difference in accuracy when compared to the group with model score, as depicted in Figure \ref{fig:metrics}.
Finally, we provide insights about the variability and agreement of the different \textit{post-hoc} explainers based on the produced explanations for the experiment.

\subsection{Data + ML Model Score}

We first analyse the difference in between human decision-making with and without presence of the ML model score. We evaluate \textit{H1} (see Section \ref{ssec:experimental_hypoehtess}) in terms of the time taken to make decisions, accuracy, recall, and \ac{FPR}, whereas \textit{H6} is examined under the agreements measures mentioned in Section \ref{ssec:metrics}.

\subsubsection{\textbf{Showing end-users the ML Model Score improves average decision time over Data Only}}

Our results show that withholding the model score leads to significantly slower decisions. Using the \textit{Mann-Whitney} \textit{U} test, we detect a significant difference in times between \textit{Data Only} and \textit{Data + ML Model Score} $\left(p<0.01\right)$ (Table \ref{tab:metrics}).  A more thorough analysis of the performance metrics (see Figure \ref{fig:times}) reveals an approximate decrease of $25\%$ of the relative average time to decide, when presenting information about the model score. 
\textbf{When considering time as the performance metric, these results corroborate} \textit{H1} (as defined in Section \ref{ssec:experimental_hypoehtess}).

\subsubsection{\textbf{Showing end-users the ML Model Score deteriorates their accuracy over Data Only}}

Our results demonstrate that withholding information about the model score significantly improves the user's predictive accuracy. 
Table \ref{tab:metrics} shows that, after the application of the \textit{Chi-squared} test, significant differences arise between \textit{Data only} and \textit{Data + ML Model Score} $\left(p=0.08\right)$. 
These results contradict \textit{H1} (when using accuracy as the performance metric).
This might derive from the fact that the instances being reviewed are in a score band near the decision threshold, and, therefore, have a higher associated uncertainty when being classified.

\subsubsection{\textbf{Showing end-users the ML Model Score does not significantly improve recall or \ac{FPR} over Data Only}}

Our results do not exhibit statistically relevant improvements in terms of other users' performance metrics like recall or \ac{FPR}. Considering these as the desired performance metric reveals to be inconclusive and, therefore, does not suffice to support nor reject \textit{H1}.

\begin{figure*}[btph]
  \includegraphics[width=\linewidth]{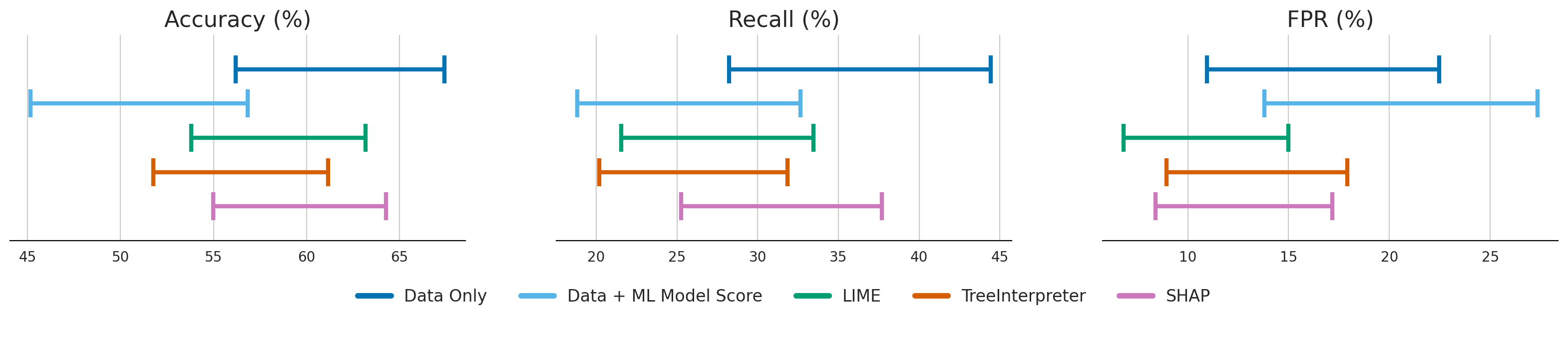}
  \caption{Confidence intervals ($90\%$) for each performance metric of all variants of the experiment. The interval is calculated through the \textit{beta distribution} for the estimated parameter $p$ of each metric.}
  \label{fig:metrics}
\end{figure*}

In general, Figure \ref{fig:metrics} shows a degradation in all metrics derived from the confusion matrix, when comparing the \textit{ML Model Score} group to \textit{Data Only}, as both recall and accuracy registered a loss of $10\%$ and \ac{FPR} registered an increase of around $4\%$ percentage points.

\subsubsection{\textbf{Showing the ML Model Score decreases agreement}}

The consensus among fraud analysts was shown to decrease as we incorporated more information. This is visible in Table \ref{tab:metrics}, as the measurement of \textit{Fleiss' Kappa} went from $0.41$ in the \textit{Data Only} variant to $-0.02$ in the \textit{Data + ML Model Score} variant. The former reflects a setting where users, on average, agreed on the transaction label $76.67\%$ of the times, whereas in the latter they only agreed on $63.33\%$ of the times. This refutes the idea that showing more information would guide (or shape) users thinking process by giving hints about relevant aspects and, consequently, disproves hypothesis \textit{H6}.

We hypothesize this large difference is due to (1) too small agreement set and (2) high proportion of transactions classified as legitimate (\textit{i.e.}, $77\%$), leading to extra sensitivity to disagreements about fraudulent transactions. 

\subsection{Data + ML Model Score + Explanations}

We further examine the performance differences between decision-making tasks involving \textit{Data Only} and \textit{Data + ML Model Score + Explanations}. In particular, we examine the impact of three distinct variants of the \textit{Data + ML Model Score + Explanations} group: LIME, SHAP, and TreeInterpreter.

\subsubsection{\textbf{Showing \textit{post-hoc} explanations significantly improves end-users average speed over Data Only}}
\begin{figure}[hbtp]
          \includegraphics[width=0.801\linewidth]{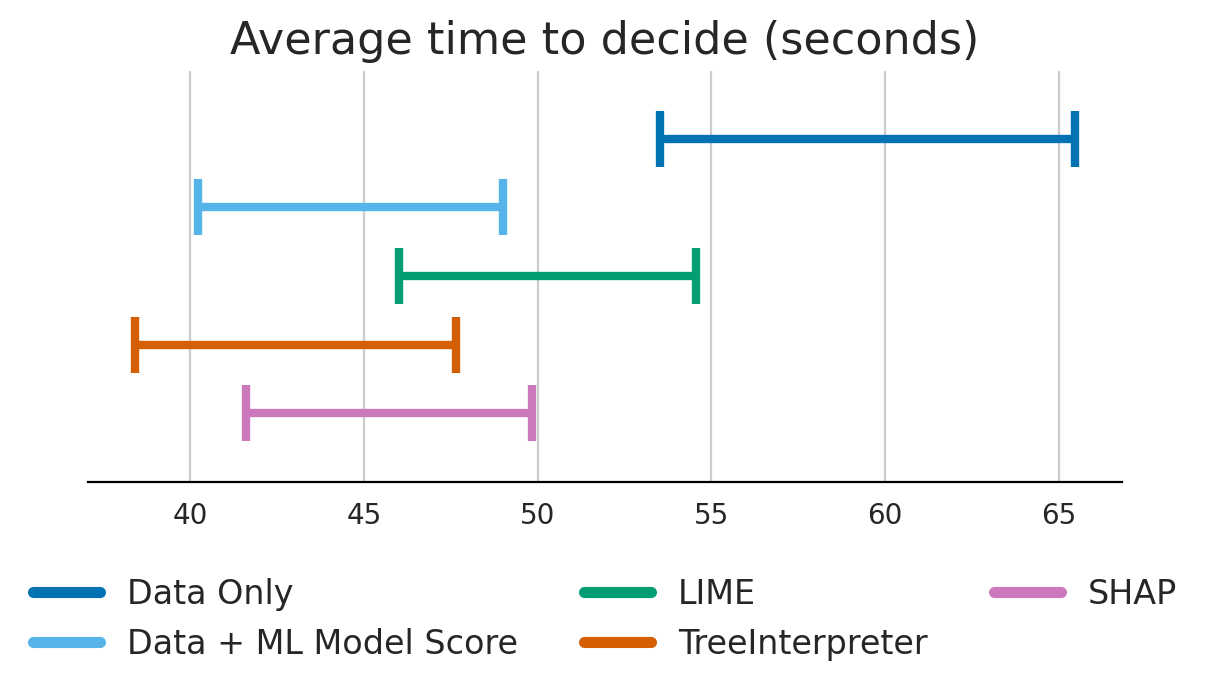}
  \caption{Confidence intervals for the average decision time of each variant. The interval represents the Standard Error of the sample multiplied by $1.64$, representing a $90\%$ Confidence Interval, centered around the mean group's mean.}
  \label{fig:times}
\end{figure}

Figure \ref{fig:times} shows the confidence intervals of decision time for each group.
By running a multiple group comparison using the \textit{Kruskal-Wallis} \textit{H} test, we observe statistically significant differences between explainer-based variants and the \textit{Data Only} group $\left(p<0.001\right)$, which corroborates \textit{H2} (when the performance metric is the reviewing time). 
We identify significant differences for every explainer, when they are compared pair-wise to the \textit{Data Only} variant by using the \textit{Holm-Bonferroni} corrected \textit{Mann-Whitney} \textit{U} tests we obtain p-values between $1\times10^{-4}$ (for LIME) and $0.09$ (for TreeInterpreter). When comparing against \textit{Data + ML Model Score}, all explainers show increased decision time but this is not statistically significant.

\subsubsection{\textbf{Different post-hoc explainers impact the end-users decision speed differently.}}
We also examine paired comparisons between the different explainers to address \textit{H3} in terms of the decision efficiency. We detect significant differences when comparing LIME to TreeInterpreter $\left(p<0.01\right)$ and SHAP to TreeInterpreter $\left(p=0.091\right)$. 
In other words, results show that among the three evaluated \textit{post-hoc} explainers, TreeInterpreter potentiates significantly faster decision-making processes. 
These results corroborate \textit{H3} when considering the average time review as the fraud analysts' measure of performance.

\subsubsection{\textbf{Showing \textit{post-hoc} explanations does not significantly improve end-users efficacy}}
In addition to efficiency, we examine the impacts of showing explanations to the human decision-making in terms of accuracy, \ac{FPR}, and recall. As visible in Figure \ref{fig:metrics}, all evaluated explainers are associated with deteriorated values for the predictive-accuracy metrics,  except for the error-based metric, \ac{FPR}. Effectively, although the values are not statistically significant, all explainers seem to lead to less false positives. 
Furthermore, as visible in Table \ref{tab:metrics} the multiple group comparison \textit{Chi-squared} test provided no conclusive results and, consequently,
no paired tests were conducted between the explainer variants. 
Notwithstanding the lower accuracy and recall values of each explainer when compared to the \textit{Data Only} variant, explainers were still able to improve upon the results obtained for the \textit{Data + ML Model Score} variant, although this improvement was also not statistically significant. 
The obtained results disprove \textit{H2} and \textit{H3}, when the considered performance metrics are either accuracy, \ac{FPR}, or recall.

Notwithstanding these results, we emphasize that, performance-wise, the selected decision time metric is the most volatile metric and, therefore, the most susceptible to vary during the experiment due to some unaccounted external factors (such as connectivity issues or distractions).

\subsubsection{\textbf{Post-hoc explainers are perceived differently in  terms of relevance, usefulness, and diversity by the end-users.}}
\begin{figure}[t]
\centering
    \subfigure{
    \includegraphics[width=.45\textwidth]{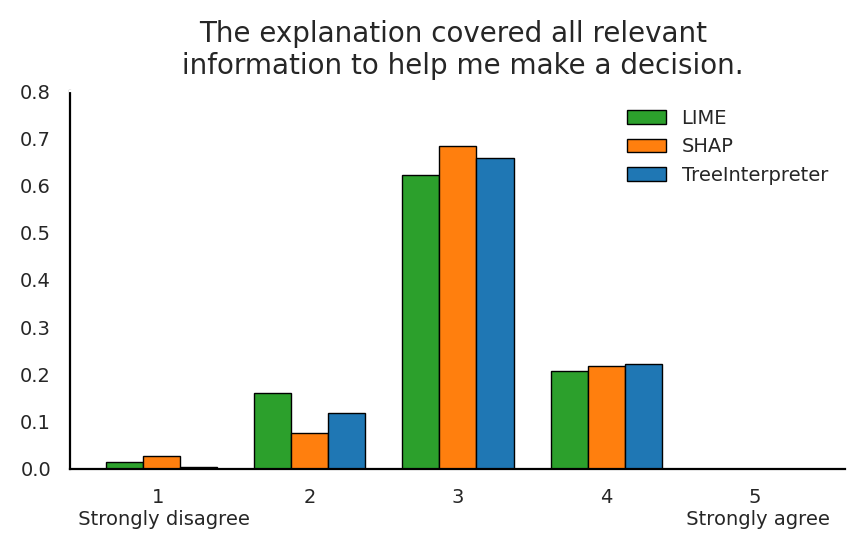}
    }
  \subfigure{
    \includegraphics[width=.45\textwidth]{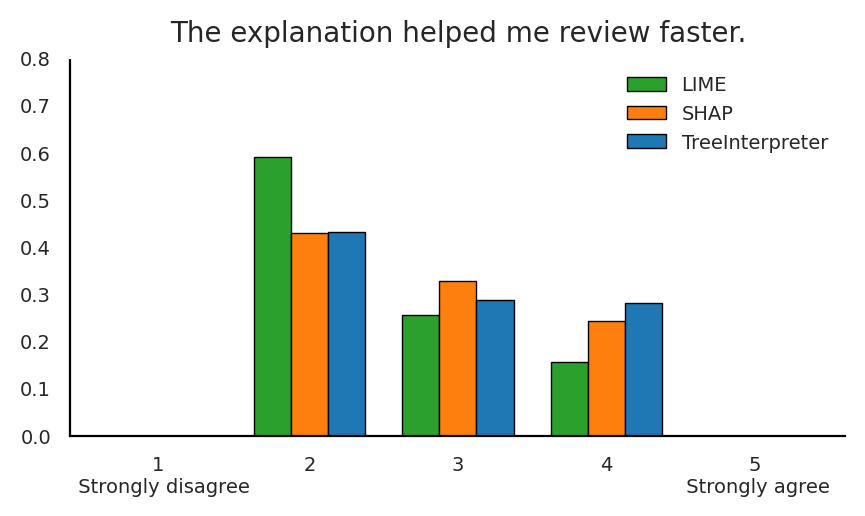}
    }
  \subfigure{
    \includegraphics[width=.45\textwidth]{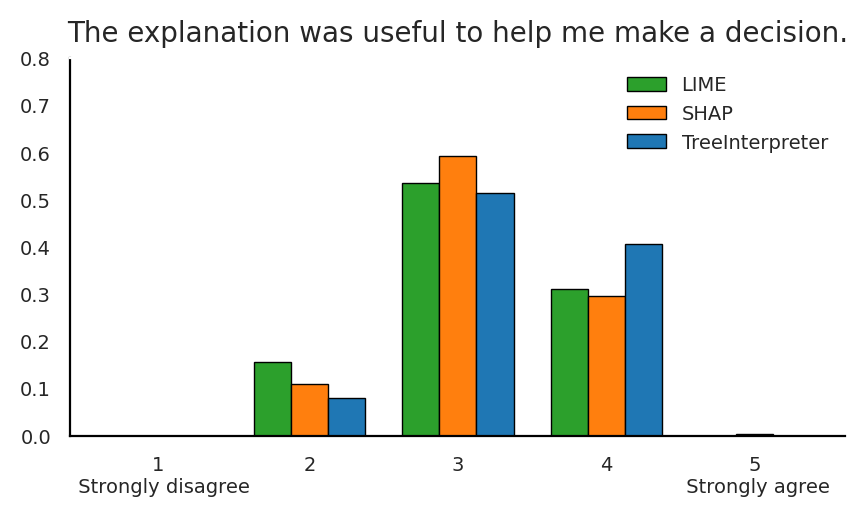}
    }
  \caption{Distribution of answers to feedback questionnaire.}
  \label{fig:answers}
\end{figure}

We perform a multiple group comparison \textit{Kruskal-Wallis} \textit{H} test to compare the results obtained with the questionnaire in Figure \ref{fig:Questionnaire}.
While no significant result is detected for the first question $\left(p=0.238\right)$, the test reveals significant changes relative to the second and third questions, that is, \textit{"The explanation helped me review faster."} $\left(p<0.001\right)$ and \textit{"The explanation was useful to help me make a decision."} $\left(p<0.01\right)$). 
Figure \ref{fig:answers} shows the distribution of the answers to the three questions posed during the last stage of the conducted experiment, discriminated by explainer. 
We observe that TreeInterpreter is the explainer with most positive answers (blue), especially in the third question. We also notice the high number of neutral answers, \textit{neither}, and practically non-existing number of extreme answers, \textit{i.e.}, \textit{strongly agree} or \textit{strongly disagree}. 
We can further observe, in statistical terms, that in the second question (middle), LIME registers a significant difference when compared to both SHAP  $\left(p<0.01\right)$ and TreeInterpreter $\left(p<0.01\right)$. On the other hand, in the third question, no paired test registered a significant difference. In this question, TreeInterpreter is the explainer with results closer to significance. 
These results support \textit{H4}, as each distinct explainers are indeed perceived differently by the users. 

\subsubsection{\textbf{Showing explanations increases end-users agreement over the same set of transactions.}}
We also examine the impacts in the agreement of the fraud analysts' decisions. 
Table \ref{tab:metrics} shows LIME to be the only explainer capable of improving the agreement beyond the \textit{Data Only} group.
However, when compared with the \textit{Data + ML Model Score} variant, all explainer variants seem to evoke more consensus among the fraud analysts. 
Quantitatively speaking, LIME achieves by far the best agreement result with a \textit{Fleiss' Kappa} of $0.53$, and fraud analysts agree, on average, on $84.62\%$ of the decisions. 
Also promising, but still inferior to the agreement achieved when all information is withheld from the user, is TreeInterpreter with a \textit{Fleiss' Kappa} of $0.30$, and with an average agreement of $69.23\%$. 
Lastly, SHAP exhibits the lowest value of agreement, with a \textit{Fleiss' Kappa} of $0.15$, an average agreement of $64.10\%$. 
These results partially corroborate \textit{H5}, as LIME actually seems to improve analysts agreement. However, the same does not verify for the other explanation methods.

\subsection{Variability in Explanations} 

We analyse the variety and agreement of the explanations used during the third stage of the experiment. To this end, we collect the explanations of the different evaluated explainers (LIME, SHAP, and TreeInterpreter) for every transaction of the experiment. Each explanation comprises six feature-contribution pairs which are the basis of the explanations. 

To better comprehend the explainers' behavior, we measured the diversity of their explanations. This implies comparing how many of the $111$ available features are actually being used to create the explanations: LIME showed the least diversity, using $34$ features ($30.6\%$ of the total set of features), followed by SHAP, which used a total of $89$ features ($80.2\%$ of the total set of features), and TreeInterpreter, which used a total of $107$ features ($96.4\%$ of the total set of features). 
Lower values in the number of used features translates into less variability in the explanations. This also ends up reflecting on the occurrence rate of the most popular feature (\textit{i.e.}, the feature used the most times to explain an instance), which in LIME occurred in $89.7\%$ of the transactions, as opposed to the most common feature in TreeInterpreter which only occurred $45.3\%$ of the explanations.

The agreement between explainers is calculated by how many features two given explainers choose to integrate the explanation normalized by the length of the explanation. For example, if in an instance LIME and SHAP had chosen 2 features in common to explain the instance score, and the other 4 features were different for each explainer, the agreement in that instance would be $33.3\%$ for that pair of explainers.

Comparing explanations between SHAP and TreeInterpreter produces an agreement of $53.0\%$, \textit{i.e.}, $53.0\%$ of the features used by SHAP for a given explanation were also used by TreeInterpreter. Likewise, the agreement for the other explainers' pairs produces an agreement of $41.0\%$ (between LIME and SHAP) and $23.5\%$ (between LIME and TreeInterpreter). These results show that the output explanation for a given instance depends on the \textit{post-hoc} method chosen to explain it, \textit{i.e.}, different explainers will choose different features to explain a given instance. 

\subsection{Study Limitations}

In this section, we outline the main limitations of our empirical study. We have a constraint in the number of participants as well as their availability for the experiment, which in turn limits the sample size for the experiment. This has an impact on the effect size, or the rates of errors for the statistical tests. To perform tests with higher sensibility to smaller changes on the measured metrics, it is necessary to increase the sample size.

Another limitation of the study is that we cannot control all the possible external factors, such as difficulty of the instance, user attention to the tested information (data, model score, and explanations), connectivity speed, among other factors. However, the mitigation of the effects of such unaccountable factors is only possible when running large scale randomized controlled trials.


This study showed no significant differences in performance metrics derived from the confusion matrix between LIME, SHAP, and TreeInterpreter, using the same explanation format. 
A relevant study is to explore how different configurations and visualizations alter the observed results.

\section{Conclusion}
\label{sec:conclusions}

The recent developments of XAI methods has not been accompanied by a robust and practical assessment of their true impact on decision-making tasks. 
More often than not, the quality of these methods is measured through proxy desiderata (\textit{e.g.}, fidelity or robustness), hence, failing to convey the information of the actual impact on the end-users' performance (\textit{e.g.}, accuracy or decision time). The lack of awareness towards the performance of the whole model + explanations + end-users may result in sub-optimal decision processes.

With this work, we hope to fill in this gap by proposing XAI Test, an application-grounded evaluation methodology suited for detaching the true impact of different information levels (\textit{e.g.}, model score, explanations) in Human-AI collaborative systems. Following XAI Test, we conducted a user study to evaluate three well-known \textit{post-hoc} explainability methods (i.e., LIME, SHAP, TreeInterpreter) on a real-world fraud detection task, encompassing 3 fraud analysts, an ML production model, and real-world data. Throughout the experiment, we progressively elevate the level of information presented to the analysts in three stages. We begin with information exclusively about the data (\textit{Data only}) and subsequentially unveil information about the ML model score (\textit{Data + ML Model Score}) and, in the last stage, about the explanations (\textit{Data + ML Model Score + Explanations}). In the course of the experiment, we collect measures of the performance of the analysts in function of the revealed information. These include the duration, the accuracy, recall, and FPR of the decisions made, as well as the user's feedback on the perceived utility of the explanations.

To the best of our knowledge, this is the first study to perform a quantitative benchmark of the impact of different explanation methods on human decision-making performance on a real-world setting (real task, real data, real users). We complement this analysis with a strong battery of statistical tests to strengthen the validity of our conclusions. Obtained results reveal that, when provided with \textit{Data only} information, fraud analysts decide significantly better but also more slowly when compared to variants that include information about the ML model. In this regard, our results show explanations (\textit{Data + ML Model Score + Explanations}) to slightly improve the accuracy upon the \textit{Data + ML Model Score} but to still fall short of the accuracy achieved in the \textit{Data only} setup. Finally, amongst the three evaluated explainers, the analysts identify LIME as the least-favoured explanation method, potentially, due to its low explanations diversity. 

In general, our results seem to suggest an existing trade-off between effectiveness and efficiency as the analysts are provided with added levels of information. This raises awareness towards blindly selecting popular \textit{post-hoc} explanation methods in real-world decision-making settings. 

\section{Acknowledgements}

The project CAMELOT (reference POCI-01-0247-FEDER-045915) leading to this work is co-financed by the ERDF - European Regional Development Fund through the Operational Program for Competitiveness and Internationalisation - COMPETE 2020, the North Portugal Regional Operational Program - NORTE 2020 and by the Portuguese Foundation for Science and Technology - FCT under the CMU Portugal international partnership.

\bibliographystyle{unsrt}
\bibliography{refs}


\end{document}